%% file: bare_conf.tex
\documentclass[conference]{IEEEtran}
\ifCLASSINFOpdf
\else
\fi
\usepackage{algorithm}
\usepackage{algorithmic}

\usepackage{xcolor}

\usepackage{graphicx}
\usepackage{amsmath}
\usepackage{amsfonts}
\usepackage{multirow}
\newcommand{\networkName}{StreamTinyNet }
\newcommand{\networkNameNoSpace}{StreamTinyNet}

\begin{document}
\title{\networkNameNoSpace: video streaming analysis with spatial-temporal TinyML}

% \author{\IEEEauthorblockN{Anonymous Authors}}
\author{\IEEEauthorblockN{Hazem Hesham Yousef Shalby}
\IEEEauthorblockA{Politecnico di Milano\\
Milan, Italy\\
hazemhesham.shalby@polimi.it}
\and
\IEEEauthorblockN{Massimo Pavan}
\IEEEauthorblockA{Politecnico di Milano\\
Milan, Italy\\
massimo.pavan@polimi.it}
\and
\IEEEauthorblockN{Manuel Roveri}
\IEEEauthorblockA{Politecnico di Milano\\
Milan, Italy\\
manuel.roveri@polimi.it}}


%Telephone: (800) 555--1212, Fax: (888) 555--1212}


%\IEEEpeerreviewmaketitle 
\maketitle

\begin{abstract}
Tiny Machine Learning (TinyML) is a branch of Machine Learning (ML) that constitutes a bridge between the ML world and the embedded system ecosystem (i.e., Internet-of-Things devices, embedded devices, and edge computing units), enabling the execution of ML algorithms on devices constrained in terms of memory, computational capabilities, and power consumption. Video Streaming Analysis (VSA), one of the most interesting tasks of TinyML, consists in scanning a sequence of frames in a streaming manner, with the goal of identifying interesting patterns. Given the strict constraints of these tiny devices, all the current solutions rely on performing a frame-by-frame analysis, hence not exploiting the temporal component in the stream of data. In this paper, we present \textit{\networkNameNoSpace}, the first TinyML architecture to perform multiple-frame VSA, enabling a variety of use cases that requires spatial-temporal analysis that were previously impossible to be carried out at a TinyML level. Experimental results on public-available datasets show the effectiveness and efficiency of the proposed solution. Finally, \textit{\networkName} has been ported and tested on the Arduino Nicla Vision, showing the feasibility of what proposed.
\end{abstract}


% \fi
\IEEEpeerreviewmaketitle

%-----------------------------------------------------------------------------
% INTRODUCTION
%-----------------------------------------------------------------------------
\section{Introduction}
\label{sec:introduction}
Tiny Machine Learning (TinyML) \cite{tiny_org} is an increasingly popular field of study that combines Machine Learning (ML) %, Tiny pervasive devices, sensors embedded systems, 
and Embedded and IoT devices, characterized by strict constraints in terms of memory (on-device available RAM is usually less than 1MB), computational power (Microcontrollers frequencies $<$ 500 KHz), and power consumption ($<$ 100mW). It has recently gained significant attention thanks to the ability to process data directly where they have been acquired, thereby enhancing privacy and security, reducing latency, improving real-time responsiveness, and being able to operate offline (i.e., without requiring a constant internet connection). \cite{buyya_is_2023,banbury_benchmarking_2021}

One area of focus within TinyML is Video Streaming Analysis (VSA), which involves scanning a sequence of frames (i.e., a video) in a streaming manner to identify interesting patterns \cite{anjum_video_2019}. However, due to technological constraints, the execution of ML models for on-device VSA is currently limited to a frame-by-frame inspection. A review of the related literature is provided in Section \ref{sec:related_works}. % This means that each frame is analyzed independently. 
Remarkably, the limitation of processing videos in a frame-by-frame manner hinders the evolution of the scene over time, hence reducing the ability of TinyML models to recognize temporal patterns.

The aim of this paper is to %answer the following question: \textit{How is it possible to design TinyML solutions able to support a multiple-frame VSA on tiny devices technologically constrained on memory, computation, and energy?}
%To tackle this challenge, this paper 
present, for the first time in the literature, a novel neural network architecture,  called \textit{\networkNameNoSpace}, %that is specifically designed for TinyML applications 
which is able to support multiple-frame VSA on tiny devices. The proposed architecture shows a significant reduction in the memory and computational requirements when compared to standard, non-tiny architectures and, at the same time, when compared to single-frame solutions present in the TinyML literature, it shows great accuracy improvements while keeping the differences in memory and computational demands small to negligible. Furthermore, the use of \textit{\networkName} enables, for the first time at a TinyML level, a variety of use cases that require spatial-temporal analysis (e.g., gesture recognition) that are impossible to be carried out with single-frame solutions.
%Specifically, the experiments have been performed on public-available datasets.
Experiments conducted on a resource-constrained device (Arduino Nicla vision\cite{noauthor_nicla_nodate}) demonstrate the feasibility of porting the architecture on real-world tiny devices.

The paper is organized as follows.
Section \ref{sec:related_works} provides an overview of the related literature.
Section \ref{sec:proposed_architecture} delves into the proposed architecture, its implementation, its memory and computational complexity, and its learning algorithm.
In Section \ref{sec:evaluation}, an evaluation of the proposed architecture is conducted on public-available benchmarks and datasets.
Section \ref{sec:porting} outlines the porting process on the Arduino Nicla Vision \cite{noauthor_nicla_nodate}.
Finally, Section \ref{sec:conclusions} discusses the main findings of this research and addresses future research directions.

%-----------------------------------------------------------------------------
% Related works
%-----------------------------------------------------------------------------
\section{Related works}\label{sec:related_works}
This section describes an overview of the related literature by organizing the works into three main topics: TinyML solutions and algorithms, VSA, and Video Classification.
\subsection{TinyML}
TinyML solutions present in the literature rely on techniques to reduce the size and complexity of the ML model. This allows the memory and computational demand to be significantly reduced at the expense of a (possibly negligible) reduction in accuracy of the model.
The techniques present in this field can be grouped into two main families: approximate computing mechanisms, and network architecture redesigning.

\subsubsection{Approximate computing mechanisms} These mechanisms, whose goal is to trade off accuracy with computational and memory demand \cite{buyya_is_2023},  can be further grouped into two main families:
\begin{itemize}
    \item \textit{Precision scaling} aims to reduce the memory occupation by changing the precision (i.e., the number of bits used for the representation) of the weights and feature maps; This mechanism relies on quantization \cite{nagel_white_2021}, and compression \cite{neill_overview_2020} techniques.
    
    \item \textit{Task dropping} aims to reduce the computational load and memory occupation by skipping the execution of certain tasks associated with the processing pipeline (such as structured and unstructured pruning mechanisms)\cite{buyya_is_2023}.
\end{itemize}

\subsubsection{Network architecture redesigning} 
Most of the research in this area focused on neural networks, specifically Convolutional Neural Networks (CNNs), since the primary frameworks available for TinyML are designed for this type of neural network.
In the literature, two common techniques are used to redesign 2D convolutions to reduce the memory and computational requirements of the CNN: Separable convolutions \cite{chollet_xception_2017}, and Dilated convolutions\cite{yu_multi-scale_2016}. The proposed \textit{\networkNameNoSpace} extends these techniques.
 
Recently, researchers in the TinyML field are increasingly focusing their interest on developing applications that rely on non-conventional sensors such as UWB (Ultra-Wideband) \cite{pavan2022device} and TOF (Time-of-Flight) \cite{noauthor_vl53l5cx_nodate,foix_lock-time--flight_2011}. 
%For instance, UWB radar modules can be used for presence detection, as suggested in \cite{pavan_tinyml_2022}. Similarly, ToF \cite{noauthor_vl53l5cx_nodate,foix_lock-time--flight_2011} can serve as a camera replacement and offers promising possibilities for various applications.

\subsection{Video stream analysis in TinyML}
In the TinyML field, analyzing video streams typically involves the deployment of tiny ML models to perform real-time analysis of video data directly on tiny devices. Examples of tasks belonging to this activity are object detection for surveillance cameras and facial recognition to identify individuals in a video stream. 

The most used architectures in the field are
MobileNetV1 \cite{howard_mobilenets_2017}, MobileNetV2 \cite{sandler_mobilenetv2_2019},
MicroNets \cite{banbury_micronets_2021},
and MCUNet \cite{lin_mcunet_2020}.
Moreover, in literature, some implementations of VSA on constrained devices are studied in 
\cite{chowdhery_visual_2019, sudharsan_tinyml-cam_2022,probierz_application_2022, tata_real-time_2022}. 
However, all the reported implementations present a limit to their application on tiny devices, which is the usage of frame-by-frame analysis, where the sequence of frames preceding the one taken into consideration is not explored.

\subsection{Video classification}
Video classification is the task that maps a sequence of frames (i.e., a video) into a pre-defined set of classes. This task aims at analyzing the content of the video to identify patterns and features that can be associated with a specific category among the available ones.
Video classification (and in general video understanding) is one of the main areas in computer vision and has been studied for decades.

In recent years, ML techniques have gained popularity due to their impressive performance and they are nowadays the most widely adopted techniques for video classification. Examples in this field include deep neural networks, CNNs, and recurrent neural networks (RNNs)\cite{peng_two-stream_2019}.

Currently, the main idea behind the most used ML methods involves combining spatial and temporal information. %In what follows, some of the most interesting and used approaches will be presented.
Many approaches rely on CNNs to extract features (e.g., Mobilenet \cite{howard_mobilenets_2017}) from individual frames and then integrate them into a fixed-size descriptor using pooling, high-dimensional feature encoding, or recurrent neural networks \cite{mcnally_golfdb_2019}.
Other CNN-based approaches follow the two-stream framework \cite{peng_two-stream_2019, simonyan_two-stream_2014}, which consists in analyzing a spatial stream that operates on individual frames and a temporal stream that operates on optical flow images, and the C3D networks \cite{tran_learning_2015,ji_3d_2013}, which consists of a series of 3D convolutional layers followed by fully connected layers.
Nevertheless, the aforementioned approaches require memory and computational demands that are far beyond the technological constraints of tiny devices.
In an effort to tackle the challenges posed by the use of 3D convolutional architecture, \cite{tran_closer_2018} presents a solution where the actor factorizes the 3D convolutional filters into "(2+1)D" distinct spatial and temporal components. This approach has proven to significantly enhance accuracy while reducing both memory usage and computational load. Despite that, implementing such an approach on TinyML devices still presents two big challenges: the necessity of storing the entire frames used for the prediction, and repeated computations over frames. Consequently, the "(2+1)D" approach is considered impractical for TinyML applications. 

Dealing with the mentioned obstacles is a fundamental step to enable multi-frame video analysis in TinyML, and our solution proposes for the first time in the literature a way to address this problem.
%Nevertheless, the proposed solution in this research effectively overcomes these obstacles, enabling multiple-frame video streaming tasks on tiny devices for the first time in literature. 

%-----------------------------------------------------------------------------
%PROPOSED SOLUTION
%---------------------------------------------------------------------------
\section{The proposed solution} \label{sec:proposed_architecture}

\begin{figure*}
    \centering
    \includegraphics[width=\textwidth]{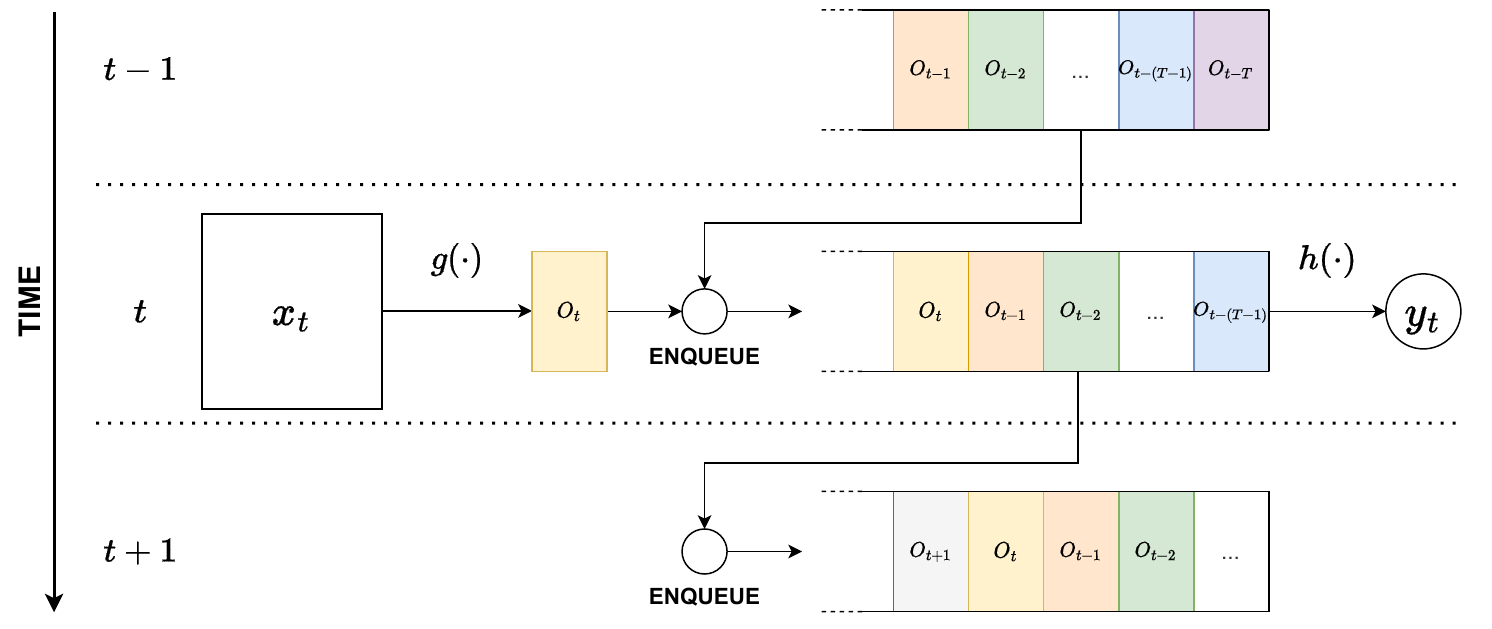}
    \caption{An overview of the proposed architecture}
    \label{fig:abstract_schema}
\end{figure*}

This section delves into the research findings and provides a comprehensive discussion of the proposed architecture for VSA on tiny devices. Specifically, Section \ref{sec:problem_formulation} presents a formulation of the problem being addressed. In Section \ref{sec:overview}, an overview of the proposed solution is provided, while in Section \ref{sec:proposed_network_sub} \textit{\networkName}is discussed in detail.
% The training process of the proposed network is discussed in Section \ref{sec:training_process}.
Finally, Section \ref{sec:network_complexity} examines the memory footprint and computational load of the proposed solution.

\subsection{Problem formulation}\label{sec:problem_formulation}
The objective of this research is to propose a neural network architecture specifically designed for VSA. Among the various tasks in video analysis, this research specifically focuses on classification. In particular, the classification is performed on a continuous basis, considering a window of the most recent $T$ frames that are currently being streamed, being $T$ an application-specific parameter, that can be tuned by the designer.

More formally, this problem can be reformulated as the design of a classifier $f_\theta (x_t, x_{t-1}, x_{t-2}, ...,x_{t-(T-1)})$ able to map the previously-unseen batch of frames $(x_t, x_{t-1}, x_{t-2}, ...,x_{t-(T-1)})$ to its label $y_t$, being $T$ defined as the length of the observation window, $x_t$ the frame acquired at time $t$ with dimensions $M_{in}\times N_{in}\times C_{in}$, and $y_t$ a label that belongs to the label set $\Omega = \{\Omega_1, \Omega_2, ..., \Omega_k \}$, where $k$ is the total number of classes related to a specific problem.

\subsection{Architecture overview}\label{sec:overview}
The proposed solution, whose graphical description is given in Fig. \ref{fig:abstract_schema}, enforces the separation between the spatial and temporal aspects within the novel network architecture. 
This approach, which is crucial in minimizing computational redundancies when analyzing frames for multiple predictions, relies on two subsequent steps: 
\begin{itemize}
    \item The spatial frame-by-frame feature extraction $g(\cdot)$;
    \item The temporal combination of the extracted features $h(\cdot)$. 
\end{itemize}

In more detail, the first step $g(\cdot)$ of the architecture serves as a feature extractor aiming at extracting information in a frame-by-frame manner within a sequence and reducing the frame dimension. We refer to this initial step as the function $g(\cdot)$, defined as:

\begin{align*}
  g(\cdot) \colon \mathbb{R}^{M_{in} \times N_{in} \times C_{in}} &\to \mathbb{R}^{M_{out} \times N_{out}  \times C_{out}}\\
  x_t &\mapsto O_t\\
\end{align*}

where:
\begin{itemize}
    \item $x_t$ and $O_t$ represent respectively the input (i.e., one frame) and the output (i.e., a feature map) of the first step at time t;
    \item $M_{in}$, $N_{in}$, and $C_{in}$ represent the input image dimension (i.e., height, width, channels of the input);
    \item $M_{out}$, $N_{out}$, and $C_{out}$ represent the output dimension (i.e., height, width, channels of the output);
    \item $g(\cdot)$ parameterized with $\theta_g$, represents the function that maps the input frame $x_t$ to the output $O_t$.
\end{itemize}

It is essential to design $g(\cdot)$ to satisfy the following inequality: 
\[M_{in}\times N_{in} \times C_{in} \gg M_{out} \times N_{out} \times C_{out}\]
so as to enforce the reduction of dimensionality between the input and output.

In the second step, the outputs of $g(\cdot)$ are jointly analyzed to exploit the temporal aspect. To be more precise, we refer to this second step as the function $h(\cdot)$, which can be defined as:
\begin{align*}
  h(\cdot) \colon \mathbb{R}^{M_{out} \times N_{out} \times C_{out} \times T} &\to \Omega \\
  O_t^T &\mapsto y_t.
\end{align*}
where:
\begin{itemize}
    \item $O_t^T \in \mathbb{R}^{M_{out} \times N_{out} \times C_{out} \times T}$   
    \item $h(\cdot)$ parameterized with $\theta_h$, represents the function that maps the outputs $O_t^T$ of the previous step to a label  $y_t \in {\Omega}$ associated to $(x_t, x_{t-1}, x_{t-2}, ..., x_{t-(T-1)})$.
\end{itemize}

\subsection{\networkName description}\label{sec:proposed_network_sub}

\begin{figure}
    \centering
    \includegraphics[width=\linewidth]{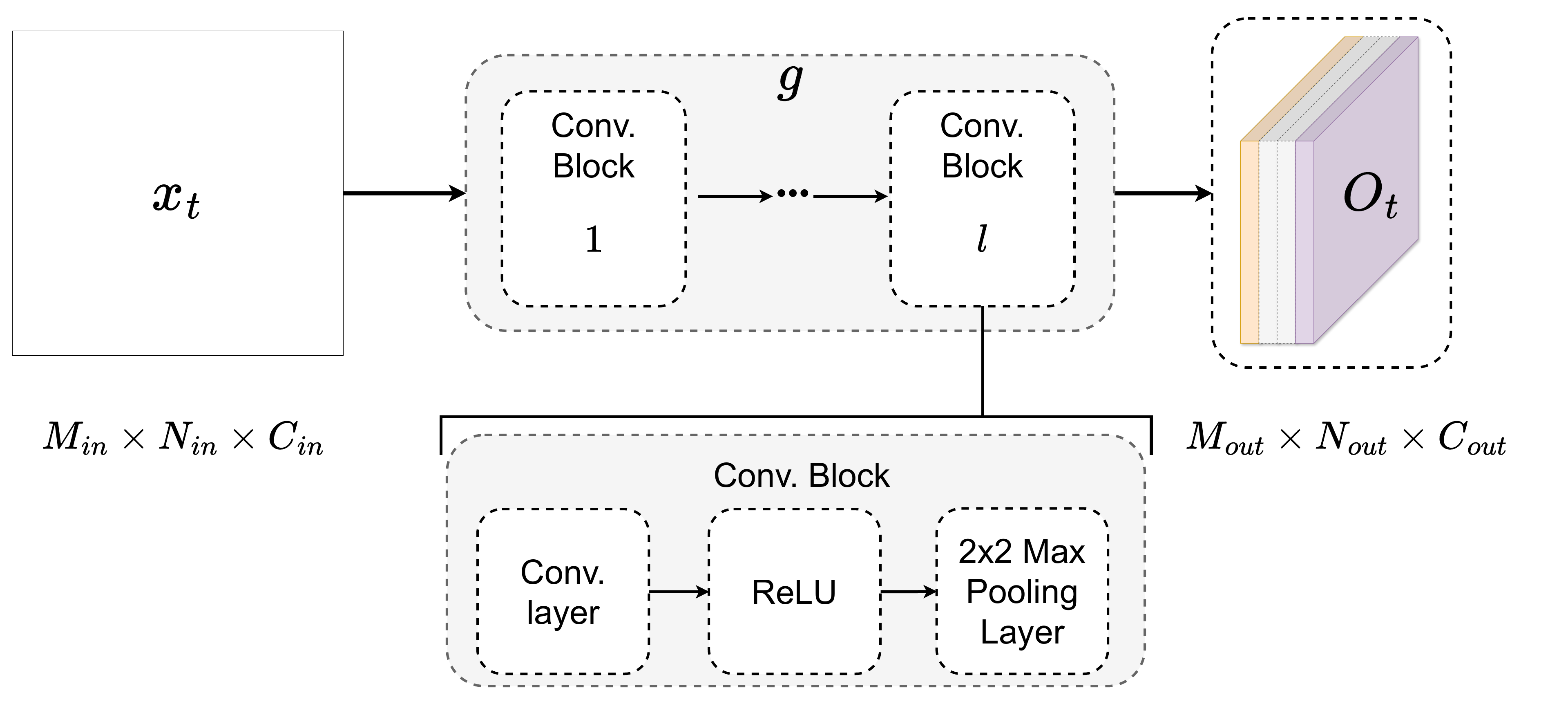}
    \caption{Representation of the feature extractor $g(\cdot)$.}
    \label{fig:feature_extractor}
\end{figure}

The aim of this section is to tailor the general architecture described in Section \ref{sec:overview} to the proposed \textit{\networkName} solution implementing the novel spatial-temporal processing for TinyML. 
In more detail, the proposed \textit{\networkName} implements the function $g(\cdot)$ by using a convolutional feature extractor, as CNNs are revealed to be the cutting-edge solution in several image-processing applications. The selected feature extractor, depicted in Figure \ref{fig:feature_extractor}, comprises $l$ sequential convolutional blocks. The $i-th$ block consists of a $r^{(i)} \times r^{(i)}$ 2D convolutional layer with $n^{(i)}$ filters, followed by a $2\times 2$ Max Pooling layer.

\begin{figure*}
    \centering
    \includegraphics[width=\linewidth]{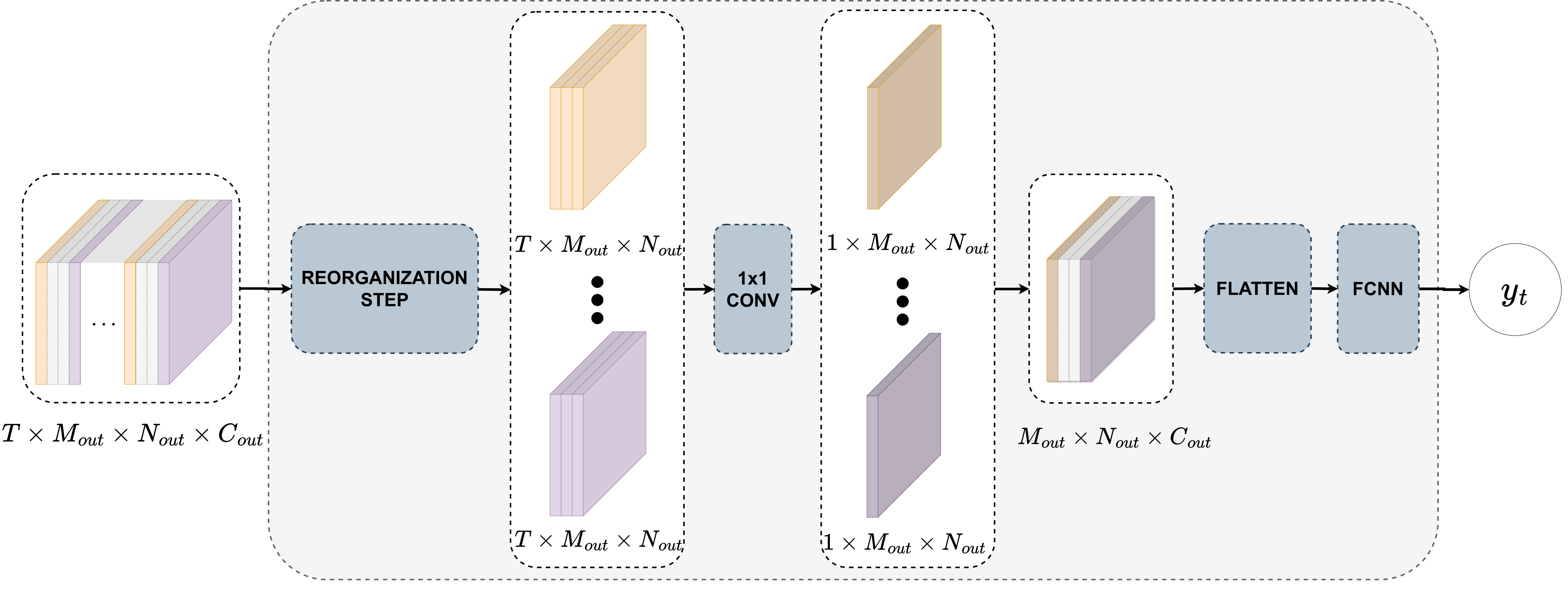}
    \caption{Representation of the proposed pipeline of $h(\cdot)$.}
    \label{fig:abstract_network_h}
\end{figure*}

Then, the function $h(\cdot)$ is implemented by using a three-step pipeline, which is reported in Figure \ref{fig:abstract_network_h}.

The first step of the pipeline uses the $T$ feature maps obtained by applying $g(\cdot)$ to the last previously-acquired $T$ consecutive frames. These maps are split along the channel axis (i.e., $C_{out}$), resulting in $C_{out}$ frames of dimension $M_{out}\times N_{out}\times 1$ for each of the $T$ feature maps. Afterward, the resulting outputs are combined into $C_{out}$ feature maps, which are obtained by stacking together the maps that belong to the same filter.

In the second step of the pipeline, $C_{out}$ different $1\times 1$ convolutions are applied to the feature maps obtained in the previous step. This process emphasizes the temporal aspect of the classification task. The output of the second step is then flattened and provided as input to the last step of the pipeline, which is a fully connected neural network (FCNN) composed of $b$ dense layer, followed by a softmax dense layer with $k$ outputs. The $j-th$ dense layer have $d^{(j)}$ units.

Summing up, Table \ref{table:parameter_summary} provides an overview of the parameters of \textit{\networkNameNoSpace}, along with their corresponding descriptions.

The proposed \textit{\networkName} is trained using an end-to-end approach, with $g(\cdot)$ and $h(\cdot)$ learned simultaneously as part of a unified training process.\footnote{ Alternatively, a pipelined training approach is also viable, where $g(\cdot)$ and $h(\cdot)$ are trained separately in a sequential manner. This may be particularly interesting in situations in which a pre-trained $g(\cdot)$ is available, resulting in reduced training times.}

\begin{table}
\caption{Overview of the parameters used to define \textit{\networkNameNoSpace}}
    \label{table:parameter_summary}
    \centering 
    \begin{tabular}{|c|c|}
    \hline
      \textbf{Parameter} & \textbf{description}\\
     \hline\hline
        $T$ & observation window\\
    \hline \hline
        $l$ &number of conv. blocks\\
        $n^{(i)}$ & number of filters of the $i-th$ conv. block \\
        $r^{(i)}$ & filters dimension of the $i-th$ conv. block\\
    \hline\hline
        $b$ & number of dense layer\\
         $d^{(j)}$ & number of units of the $j-th$ dense layer\\
         $k$ & number of output classes\\
    \hline
    \end{tabular}
     
\end{table}

\input{Tab_complexity_2}

\subsection{Network complexity}\label{sec:network_complexity}

This section introduces an analytical evaluation of the computational load $c$ and memory footprint $m$ of proposed \textit{\networkNameNoSpace}, which is essential to support the porting phase on a tiny device characterized by constraints on memory $\Bar{m}$ and computation $\Bar{c}$. In particular, $m$ and $c$ are defined as follows:
\begin{align*}
    & m = m_w + m_a < \Bar{m}\\
    & c = \sum_{l \in f} c(l) < \Bar{c}%c^{g} + c^{h} 
\end{align*}
where:
\begin{align*}
    & m_w = \sum_{l \in f} m_w(l) \\ %m_w^g + m_w^h \\
    & m_a = m_a^g + m_a^h 
\end{align*}
being:
\begin{itemize}
    \item $c(l)$ the computational load of the layer $l$ of the network. 
%and $c^h$ the computation load required for executing functions $g$ and $h$ respectively.
    \item $m_w(l)$ the total number of parameters of layer $l$.
    \item $m_a^g$ and $m_a^h$ the amount of values required to store the activations of $g$ and $h$, respectively.
\end{itemize}

To compute $m_a^g$ and $m_a^h$, we have considered the optimizations in \cite{pavan_tinyml_2022}, so that $m_a^g$ and $m_a^h$ are equal to the maximum sum of the memory required for the activations of two consecutive layers of $g$ and $h$, respectively.

Building upon the formalism established in \cite{buyya_is_2023} and \cite{pavan_tinyml_2022}, the memory demand and the computational load of each sub-layer $l$ of the network are computed as detailed\footnote{In the memory demand in Table \ref{table:complexity_g} the biases are omitted for simplicity} in Table \ref{table:complexity_g}.

More in detail, when a dense layer with $d$ dense unit is applied to a tensor of dimension $N_{in}$, the memory load for the weights $m_w$, the memory for the activations $m_a$, and the computational load $c$ are defined as follows:
\begin{align*}
    m_{w} = N_{in} \times d + d&\text{,}\\
    m_{a} = d&\text{,}\\
    c = d \times N_{in}&\text{.}
\end{align*}
Differently, when a Convolutional Layer with $n$ filters of dimension $r \times r$ is applied to an input with dimension $M_{in} \times N_{in} \times n_{in}$, the memory for the weights $m_w$, the memory for the activations $m_a$, and computational load $c$ are defined as follows:
\begin{align*}
    % m^{CONV} = m^{CONV}_{w} + m^{CONV}_{a} \\
    m_{w} = r \times r \times n_{in} \times n + n &\text{,}\\
    m_{a} = M_{in} \times N_{in} \times n&\text{,}\\
    c = r \times r \times n_{in}\times n \times M_{in}\times N_{in}&\text{.}
\end{align*}
A special case of convolutions in the proposed architecture is the $1\times1$ CONV within $h(\cdot)$, which is always applied to an input of dimension $M_{out}\times N_{out}\times C_{out}\times T$, i.e., to the outputs of $g(\cdot)$ in the observation window $T$. Therefore, the storage of the $T$ feature maps generated by $g(\cdot)$ requires an increase of $M_{out}\times N_{out}\times C_{out}\times T$ to the usual activations memory of convolutions.
It is noteworthy to point out that when two consecutive observation windows $O_t^T$ and $O_{t+1}^T$ do not overlap, the $1\times 1$ convolution step within $h(\cdot)$ can be computed in a streaming manner. This eliminates the need to store the $T$ feature maps generated by $g(\cdot),$ thereby resulting in a further reduction in the memory required for inference.

We also emphasize that the memory footprints reported in Table \ref{table:complexity_g} are measured in terms of the number of weights ($m_w$) and the number of values required to store the activations ($m_a$), and thus, to obtain actual memory requirements, they should be multiplied by the dimension in Bytes of the format of those numbers (i.e., 1B for 8bit-Integers, or 4B for 32bit-Floats).    
Differently, the computational load (i.e., $c$) is measured in terms of the number of floating-point operations (FLOPs).

\section{Experimental results}\label{sec:evaluation}
This section introduces the experimental results measuring the effectiveness and the efficiency of the proposed \textit{\networkName} on two challenging VSA tasks: Gesture Recognition, and Event Detection. We emphasize that both tasks can be formulated as the classification of frames within a video stream.

To assess the performance of the proposed \textit{\networkName} architecture, we considered as a comparison three existing architectures from the literature, i.e., MobileNetV1 (with $\alpha = 0.25$)\cite{howard_mobilenets_2017}, MobileNetV2(with $\alpha = 0.3$)\cite{sandler_mobilenetv2_2019}, and MCUNet \cite{lin_mcunet_2020}. These solutions are based on frame-by-frame solutions as these are currently the only implementations present in the literature within a TinyML setting. Therefore, to ease the comparison, a majority voting approach has been applied to these solutions(i.e., the most prevalent prediction among the $T$ frames is considered). All the compared architectures have been trained from scratch on the same frames within the windows used to train the proposed solution.

\subsection{Gesture recognition}\label{sec:ev_jester}
Gesture Recognition is the task of classifying which gesture is being performed by a human in front of the device, by analyzing the video stream collected by the on-device camera.
For this purpose we considered the Jester gesture recognition dataset\cite{materzynska_jester_2019}, which comprises $148,092$ labeled video clips, where humans perform fundamental hand gestures in front of a laptop camera.

\begin{table}
 \caption{Overview of the parameters of \textit{\networkName}used in the two experiments.}
     \label{table:parameter_summary-golf}
    \centering 
    \begin{tabular}{|c|c|c|}
    \hline
     \textbf{Parameter} & \textbf{Event Detection} & \textbf{Gesture Recognition}\\
     \hline\hline
     $M_{in}\times N_{in} \times C_{in}$ &$160\times160\times3$&$120\times120\times3$\\
     \hline\hline
      $T$ & 16 & 10\\
    \hline \hline
         $l$ & $5$ & $4$\\
         \hline
         $n^{(1)}$ & $4$ & $4$ \\
         $n^{(2)}$ &$8$ & $8$\\
         $n^{(3)}$ &$16$ & $16$ \\
         $n^{(4)}$ &$32$ & $32$\\
         $n^{(5)}$ &$64$ & -\\
         \hline
         $r^{(-)}$ & $2$ & $2$\\
    \hline\hline
         $b$ & 2 & 1\\
         $d^{(1)}$ & $64$ & $16$\\
         $d^{(2)}$& $32$ & - \\
         $k$ & 9 & 3\\
    \hline
    \end{tabular}
    %\\[10pt]
   
\end{table}

\subsubsection{Dataset generation}
The dataset considered for this experiment is a subset of the Jester dataset, comprising three classes: No Gesture (class $0$, $5,344$ samples), Sliding Two Fingers Down (class $1$, $5,410$ samples), and Sliding Two Fingers Up (class $2$, $5,262$ samples).
The selection of the three classes aims to highlight the importance of employing a multiple-frame VSA.
The selected dataset undergoes a preprocessing step that involves transforming each video sample in the dataset into a batch of $T$ frames. %These frames are used to train the network on the corresponding class associated with the dataset sample. 
This process is accomplished by uniformly sampling $T$ frames from each video.

\subsubsection{Evaluation}
Table \ref{table:parameter_summary-golf} details the parameters overview of the selected \textit{\networkName}model as described in Section \ref{sec:proposed_network_sub}. The model was chosen after a grid search of the parameters on the validation set.
The choice of $T=10$ has been made based on the characteristics of the Jester dataset.

Following what is described in \cite{materzynska_jester_2019}, the dataset is divided into training, validation, and testing sets following the 8:1:1 ratio.
The accuracy and the performance metrics (i.e., $c$ and $m$) of \textit{\networkName}and the comparisons are reported in Table \ref{table:gesture_results}. 

\input{Tab_comparison_gesture}

These results highlight the significance of the proposed \textit{\networkName}multi-frame approach in the gesture recognition task, emphasizing that addressing the temporal aspect is crucial compared to a frame-by-frame approach. Indeed, in contrast to the compared architectures, in which all predictions belong to the same class, our solution achieves an accuracy of $0.81$. Furthermore, the achieved performance is not coupled with higher memory usage or additional computational requirements compared to the alternative solutions. In particular, as detailed in Table \ref{table:gesture_results}, our solution achieves a computational and a memory footprint ($c$ and $m$), which are at least $4.95$ and $1.8$ times smaller than the compared architectures. In the table, the memory footprints of the comparisons are indicated with $>$, since only the memory required for storing the parameters was computed for these solutions.

\subsection{Event Detection}\label{ex_golfdb} \label{sec:rw_golfDB}

The GolfDB \cite{mcnally_golfdb_2019} dataset is a benchmark video dataset for the task of golf swing sequencing. The dataset comprises 1400 golf swing videos of male and female professional golfers. 
Golf swings have 8 distinct events that can be localized within a frame sequence.
In the following, we will use the "Percentage of Correct Events" within the tolerance ($PCE$), introduced in \cite{mcnally_golfdb_2019}, as main figure of merit to measure the correct detection ability.

\subsubsection{Dataset generation}
The golf dataset undergoes a preprocessing step where each video sample is processed to produce batches of $T$ frames.
Specifically, significant frames are extracted from each video and, for each of these frames, a batch is constructed by including the $T-1$ frames that precede it, along with the frame itself.

\subsubsection{Evaluation}
The specific \textit{\networkName}model considered in this experimental selection has been selected through a grid search exploration on the parameters detailed in Table \ref{table:parameter_summary-golf}.
The choice of $T=16$ has been made based on the limit given by the available processing capability.
In addition, during the training process, as recommended by \cite{mcnally_golfdb_2019}, random horizontal flipping and random affine transformations ($-5^\circ$ to $+5^\circ$ rotation) are applied to the input sequences.

Following the approach used in \cite{mcnally_golfdb_2019} we considered four different splits\footnote{In each split, 75\% of the data is dedicated to training, 25\% to testing} of the dataset.
The average PCE, the performance metrics of \textit{\networkNameNoSpace}, and the comparisons are reported in Table \ref{table:goolfDB_results_t15}. As before, the memory footprints of the comparisons are indicated with $>$, since only the memory required for storing the parameters was computed for these solutions.

\input{Tab_comparison2}
\input{Tab_different_T}

The results show that distinguishing the 8 events within a golf swing by frame-by-frame solutions provides PCEs in the range $0.41-0.48$.
However, employing the \textit{\networkName}multiple-frame approach proves beneficial in distinguishing notably similar
events within the swing sequence.
Notably, an enhancement of at least $8\%$ over the frame-by-frame solution is observed with $T=16$. This improvement can be further augmented by increasing the observation window size $T$. Indeed, as shown in Table \ref{table:goolfDB_increasing_T-dataset}, augmenting $T$ results in better performances without a significant increase in the requirements (i.e., memory $m$ and computation $c$).
Specifically, the solution yielding the best performance (i.e., \textit{\networkName}with $T=16$) shows only a marginal increase of $1.01$ and $1.08$ in computational $c$ and memory $m$ requirements compared to the configuration with the smallest value of $T$.

\section{Porting}\label{sec:porting}
%In Section \ref{sec:ev_jester}, the task of gesture recognition is discussed. 
This section describes the porting results of the \textit{\networkNameNoSpace} performing the gesture recognition task to the Arduino Nicla Vision \cite{noauthor_nicla_nodate}, a device commonly used for TinyML applications.
This device is equipped with an STM32H747AII6 Dual Arm Cortex M7/M4 IC Microcontroller, 2MB of Flash Memory, 1MB RAM Memory and an integrated 2 MP Color Camera sensor. 

In order to be ported on the device, 8-Bit integer post-training quantization \cite{jacob2018quantization} was applied to the model described in Section \ref{sec:ev_jester}. The quantized model suffered a small accuracy loss on the test set after quantization, achieving a final test accuracy of 0$.79$.
The inference time on the target device was $0.065s$, enough to make the application running at $15$ fps. The total RAM usage of the whole application is approximately $300$ KB. Finally, we also estimated the energy consumption of the model for a single inference. Considering an input voltage of $5V$ and a current consumption of $105 mA$ \cite{noauthor_nicla_nodate}, the energy amount required for each single inference is $34 mJ$.

%-----------------------------------------------------------------------------
% CONCLUSION
%-----------------------------------------------------------------------------
\section{Conclusions And Future Works}\label{sec:conclusions}
%\color{black}
This research focused on the task of VSA on tiny devices, contributing significantly to the field of TinyML. The proposed \textit{\networkNameNoSpace} architecture enables multiple frames VSA for the first time in the literature. Previously, this task was restricted to frame-by-frame analysis, neglecting any temporal aspects. 
%In particular, this research introduced 
%\networkName is a novel Neural Network architecture designed to efficiently perform video classification tasks on tiny devices. 
%Experimental results show that \networkName reduces significantly the memory footprint and the computational load necessary to compute VSA compared to the other architecture implementing the same task.
Experimental results show that including multiple frames in the analysis 
 can have big advantages in terms of accuracy, while at the same time introducing minimal overheads with respect to single-frame solutions.

%The research also presents \networkName which is an efficient way of implementing the proposed architecture, and has been evaluated using different real-world datasets for sequence classification. %Additionally, while the study primarily focuses on classification tasks, its findings can be extended to other applications like object detection, tracking, and recognition.

Future works will encompass the introduction of an adaptive frame rate to optimize the power consumption in static scenes\cite{gowda_smart_2020,wu_adaframe_2019, bhardwaj_efficient_2019}, the introduction of a sensor drift detection mechanism, the extension of the architecture to include Early Exits mechanisms \cite{scardapane_why_2020} and on-device incremental training of the algorithm \cite{disabato_incremental_2020,ren_tinyol_2021}.


\section*{Acknowledgment}
This work was carried out in the EssilorLuxottica Smart Eyewear Lab, a Joint Research Center between EssilorLuxottica and Politecnico di Milano.
The authors would like to thank Ing. G. Viscardi from Politecnico di Milano for the support in the project.





% \end{thebibliography}

\bibliographystyle{plain}
\bibliography{ref_custom}

\end{document}

%% file: Tab_complexity_2.tex
\begin{table*}[t]

     \caption{The memory footprint and computational load of the sublayers of $g(\cdot)$ (see Figure \ref{fig:feature_extractor}), and $h(\cdot)$ (see Figure \ref{fig:abstract_network_h})\\
     \textit{
      $m_w(l)$, $m_a(l)$, and $c(l)$ are respectively the memory needed to store the weights, the memory needed to store the activations, and the computational load of each layer $l$. The other parameters are detailed in Table \ref{table:parameter_summary}.
     }}
     % \textbf{dobbiamo definire anche qui cosa è mw, ma e cl così come gli altri simboli della tabella.}}
    \label{table:complexity_h}
    \label{table:complexity_g}
    
    \centering 
    \begin{tabular}{|c|c|c|c|c|}
    \hline
    &$l$&$m_w(l)$  & $m_a(l)$ & $c(l)$ \\
    \hline \hline
    \multirow{6}{*}{$g(\cdot)$} &$I$ & - & $M_{in}\times N_{in}\times C_{in}$ & - \\
    \cline{2-5}
    &CONV-$1$ & $ r^{(1)^2} \times C_{in} \times n^{(1)}$& $M_{in}\times N_{in}\times n^{(1)}$ & $ r^{(1)^2}\times C_{in}\times n^{(1)}\times M_{in}\times N_{in}$ \\
    &POOL-$1$ & $ 0$ & $\frac{M_{in}}{2}\times \frac{N_{in}}{2}\times n^{(1)}$ & $2\times 2\times M_{in}\times N_{in} \times n^{(1)}$ \\
    \cline{2-5}
    &\vdots &\vdots &\vdots & \vdots\\
    \cline{2-5}
    &CONV-$l$ & $ r^{(l)^2} \times n^{(l-1)} \times n^{(l)}$ & $\frac{M_{in}}{2^{l-1}}\times \frac{N_{in}}{2^{l-1}}\times n^{(l)}$ & $ r^{(l)^2}\times n^{(l-1)}\times n^{(l)}\times \frac{M_{in}}{2^{l-1}}\times \frac{N_{in}}{2^{l-1}}$ \\
    &POOL-$l$ & $0$ & $\frac{M_{in}}{2^{l-2}}\times \frac{N_{in}}{2^{l-2}}\times n^{(l)}$ & $2\times 2\times\frac{M_{in}}{2^{l-1}}\times \frac{N_{in}}{2^{l-1}} \times n^{(l)}$ \\

    \hline \hline
    \multirow{7}{*}{$h(\cdot)$} &$O^T$& - & $M_{out}\times N_{out}\times C_{out} \times T$ & - \\
    \cline{2-5}
    &CONV($1\times 1$) & $C_{out}\times T$ & $ M_{out}\times N_{out}\times C_{out}$ & $ C_{out}\times T \times M_{out}\times N_{out}$ \\
    \cline{2-5}
    &FLATTEN & - & - & - \\
    \cline{2-5}
    &DENSE-$1$ & $ M_{out}\times N_{out}\times C_{out}\times d^{(1)}$ & $d^{(1)}$ & $ M_{out}\times N_{out}\times C_{out}\times d^{(1)}$ \\
    &\vdots &\vdots &\vdots & \vdots\\
    &DENSE-$b$ & $d^{(b-1)} \times d^{(b)}$ & $d^{(b)}$ & $d^{(b-1)}\times d^{(b)}$ \\
    &SOFTMAX & $d^{(b)}\times k$  & $k$ & $d^{(b)}\times k$ \\
    \hline
    \end{tabular}
    %\\[10pt]

\end{table*}

%% file: Tab_comparison_gesture.tex
\begin{table}[h]
    \centering
    \caption{Results of the proposed \networkName($T=10$) and of the comparisons for the Jester dataset.}
    \label{table:gesture_results}
    \begin{tabular}{|c c c c|}
    \hline
    Solution & Acc. & c($10^9$OPs) & $m$ \\
     %& \textbf{Mean} & \textbf{Variance} \\
    \hline \hline
    MobileNetV1 & 0.40 & 0.0218 & $>$ 286,931 \\
    MobileNetV2 & 0.35 & 0.0386 & $>$ 1,051,619 \\
    MCUNet & 0.34 & 0.0459 & $>$ 369,131 \\

    % \networkName(T=1) & \multicolumn{3}{c|}{Not Feasible}\\
    \textbf{\networkName} & \textbf{0.81} & \textbf{0.0044} & \textbf{160,431} \\
    % Conv3D(T=10) & 0.88 & 0.0845 & 1,264,479\\
    \hline
    \end{tabular}
    
\end{table}

% m_w
%& \textbf{28,271}
%& 256,479 

%% file: Tab_comparison2.tex
\begin{table}[h]
    \centering
    \caption{Results of the proposed \networkName($T=16$) and of the comparisons for the GolfDB dataset.}
    \label{table:goolfDB_results_t15}
    \begin{tabular}{|c c c c|}
    \hline
    Solution & PCE & c($10^9$OPs) & $m$ \\
     %& \textbf{Mean} & \textbf{Variance} \\
    \hline \hline
    % \networkName(T=1) & 0.23 & 0.007983 & 299,728\\
    MobileNetV1 & 0.48 & 0.0437 & $>$ 625,113\\
    MobileNetV2 & 0.43 & 0.0663 & $>$ 2,446,569 \\
    MCUNet & 0.41 & 0.0775 & $>$ 370,097 \\
    \textbf{\networkName} & \textbf{0.56} & \textbf{0.0080} & \textbf{321,488}\\
    % SwingNet(T=16) & 0.71 & 2.70 & $>$5,380,000 \\
    \hline
    \end{tabular}
    
\end{table}

%% file: Tab_different_T.tex
\begin{table*}[t]
     \caption{Evaluation results of \networkName with different values of $T$ for the GolfDB dataset. }
    \label{table:goolfDB_increasing_T-dataset}
   \centering 
    % \begin{tabular}{|c| c c c|}
    % \hline
    % \multicolumn{4}{|c|}{\textbf{SwingNet-160}}\\
    
    % \hline\hline
    % T & PCE&c ($10^9$ FLOPs)& $m_w$ \\
    % \hline\hline
    % 4 & 0.63& 0.68& $5,380,000$\\
    % 8 & 0.66&1.35& $5,380,000$ \\
    % 16 & 0.71&2.70 & $5,380,000$ \\
    % % 32 & 0.74&5.41& $5,380,000$ \\
    % % 64 & 0.76&10.92& $5,380,000$ \\

    % \hline
    % \end{tabular}
    % \hspace{0.01 mm} 
    \begin{tabular}{|c| c c c c c|}
    \hline
    \multicolumn{6}{|c|}{\textbf{\networkNameNoSpace}}\\
    
    \hline\hline
    T & PCE& c ($10^9$ FLOPs) & $m_w$ & $m_a$ & $m$\\
    \hline\hline
    1 & 0.23&0,007983 & 115,728& 180,800& 296,528\\
    4 & 0.48&0.007992& 115,920 & 185,600 & 301,520\\
    8 & 0.51& 0,008005& 116,176 & 192,000 & 308,176\\
    16 & 0.56& 0,008031& 116,688 & 204,800 & 321,488\\

    \hline
    % \multicolumn{6}{c}{}\\
    % \multicolumn{6}{c}{}\\
    \end{tabular}

\end{table*}